\pdfoutput=1 

\documentclass[10pt,twocolumn,letterpaper]{article}

\usepackage[pagenumbers]{cvpr} 

\usepackage{graphicx}
\usepackage{amsmath}
\usepackage{amssymb}
\usepackage{booktabs}
\usepackage{multirow}
\usepackage{caption}
\usepackage{cite}
\usepackage{algorithmic}
\usepackage{algorithm}
\usepackage{arydshln}
\usepackage{colortbl}
\usepackage{xcolor}

\usepackage[pagebackref,breaklinks,colorlinks]{hyperref}

\usepackage[capitalize]{cleveref}
\crefname{section}{Sec.}{Secs.}
\Crefname{section}{Section}{Sections}
\Crefname{table}{Table}{Tables}
\crefname{table}{Tab.}{Tabs.}

\def\cvprPaperID{10276} 
\def\confName{CVPR}
\def\confYear{2023}

\begin{document}

\title{Neuro-Modulated Hebbian Learning for Fully Test-Time Adaptation}

\author{
Yushun Tang\textsuperscript{1,2}, 
Ce Zhang\textsuperscript{1}, 
Heng Xu\textsuperscript{1}, 
Shuoshuo Chen\textsuperscript{1}, \\
Jie Cheng\textsuperscript{2}, 
Luziwei Leng\textsuperscript{2}, 
Qinghai Guo\textsuperscript{2*}, 
Zhihai He\textsuperscript{1,3*}\\
{\normalsize \textsuperscript{1}Department of Electronic and Electrical Engineering, Southern University of Science and Technology, Shenzhen, China}\\ 
{\normalsize \textsuperscript{2}Advanced Computing and Storage Laboratory, Huawei Technologies Co., Ltd., Shenzhen, China}\\
{\normalsize \textsuperscript{3}Pengcheng Laboratory, Shenzhen, China}\\
{\tt\small \{tangys2022, zhangc2019, xuh2022, chenss2021\}@mail.sustech.edu.cn} \\ 
{\tt\small \{chengjie8, lengluziwei, guoqinghai\}@huawei.com, hezh@sustech.edu.cn}
}
\maketitle
\renewcommand{\thefootnote}{\fnsymbol{footnote}}
\footnotetext[1]{~Corresponding authors.}
\renewcommand{\thefootnote}{\arabic{footnote}}

\begin{abstract}
Fully test-time adaptation aims to adapt the network model based on sequential analysis of input samples during the inference stage to address the cross-domain performance degradation problem of deep neural networks. We take inspiration from the biological plausibility learning where the neuron responses are tuned based on a local synapse-change procedure and activated by competitive lateral inhibition rules. Based on these feed-forward learning rules, we design a soft Hebbian learning process which provides an unsupervised and effective mechanism for online adaptation. We observe that the performance of this feed-forward Hebbian learning for fully test-time adaptation can be significantly improved by incorporating a feedback neuro-modulation layer. It is able to fine-tune the neuron responses based on the external feedback generated by the error back-propagation from the top inference layers. This leads to our proposed neuro-modulated Hebbian learning (NHL) method for fully test-time adaptation. With the unsupervised feed-forward soft Hebbian learning being combined with a learned neuro-modulator to capture feedback from external responses, the source model can be effectively adapted during the testing process. Experimental results on benchmark datasets demonstrate that our proposed method can significantly improve the adaptation performance of network models and outperforms existing state-of-the-art methods.
\end{abstract}

\section{Introduction}
\label{sec:intro}
Although deep neural networks have achieved great success in various machine learning tasks, their performance tends to degrade significantly when there is data shift~\cite{quinonero2008dataset,krizhevsky2017imagenet} between the training data in the source domain and the testing data in the target domain \cite{mirza2021robustness}.
To address the performance degradation problem, unsupervised domain adaptation (UDA)~\cite{long2015learning,ganin2015unsupervised,pei2018multi} has been proposed to fine-tune the model parameters with a large amount of unlabeled testing data in an unsupervised manner. 
Source-free UDA methods~\cite{liang2020we,Wang_2022_CVPR,li2020model} aim to adapt the network model without the need to access the source-domain samples. 

There are two major categories of source-free UDA methods. The first category needs to access the whole test dataset on the target domain to achieve their adaptation performance \cite{liang2020we,Wang_2022_CVPR}. 
Notice that, in many practical scenarios when we deploy the network model on client devices, the network model does not have access to the whole dataset in the target domain since collecting and constructing the test dataset in the client side is very costly. The second type of method, called fully test-time adaptation, only needs access to live streams of test samples \cite{DBLP:conf/icml/SunWLMEH20,wang2020tent,mirza2022norm}, which is able to dynamically adapt the source model on the fly during the testing process. 
Existing methods for fully test-time adaptation mainly focus on constructing various loss functions to regulate the inference process and adapt the model based on error back-propagation. 
For example, the TENT method \cite{wang2020tent} updates the batch normalization module by minimizing an entropy loss. The TTT method \cite{DBLP:conf/icml/SunWLMEH20} updates the feature extractor parameters according to a self-supervised loss on a proxy learning task.  
The TTT++ method \cite{liu2021ttt++} introduces a 
feature alignment strategy based on online moment matching. 

\subsection{Challenges in Fully Test-Time UDA}
We recognize that most of the domain variations, such as changes in the visual scenes and image transformations or corruptions, are early layers of features in the semantic hierarchy \cite{wang2020tent}. They can be effectively captured and modeled by lower layers of the network model. 
From the perspective of machine learning, early representations through the lower layer play an important role to capture the posterior distribution of the underlying explanatory factors for the observed input~\cite{bengio2013representation}. For instance, in deep neural network models, the early layers of the network tend to respond to corners, edges, or colors. In contrast, deeper layers respond to more class-specific features~\cite{zeiler2014visualizing}. In the corruption test-time adaptation scenario, the class-specific features are always the same because the testing datasets are the corruption of the training domain. However, the early layers of models can be failed due to corruption. 

Therefore, the central challenge in fully test-time UDA lies in how to learn useful early layer representations of the test samples without supervision. 
Motivated by this observation, we propose to explore neurobiology-inspired Hebbian learning 
for effective early layer representation learning and fully test-time adaptation.
It has been recognized that the learning rule of  supervised end-to-end deep neural network training 
using back-propagation and the learning rules of the early front-end neural processing in neurobiology are unrelated \cite{krotov2019unsupervised}. 
In real biological neural networks, the neuron responses are tuned based on the local activities of the presynaptic cell and the postsynaptic cell, plus some global variables to measure how well the task was carried out, but not on the activities of other specific neurons \cite{whittington2019theories}. 

\begin{figure}[!ht]
\centering
\includegraphics[width = 0.5\textwidth]{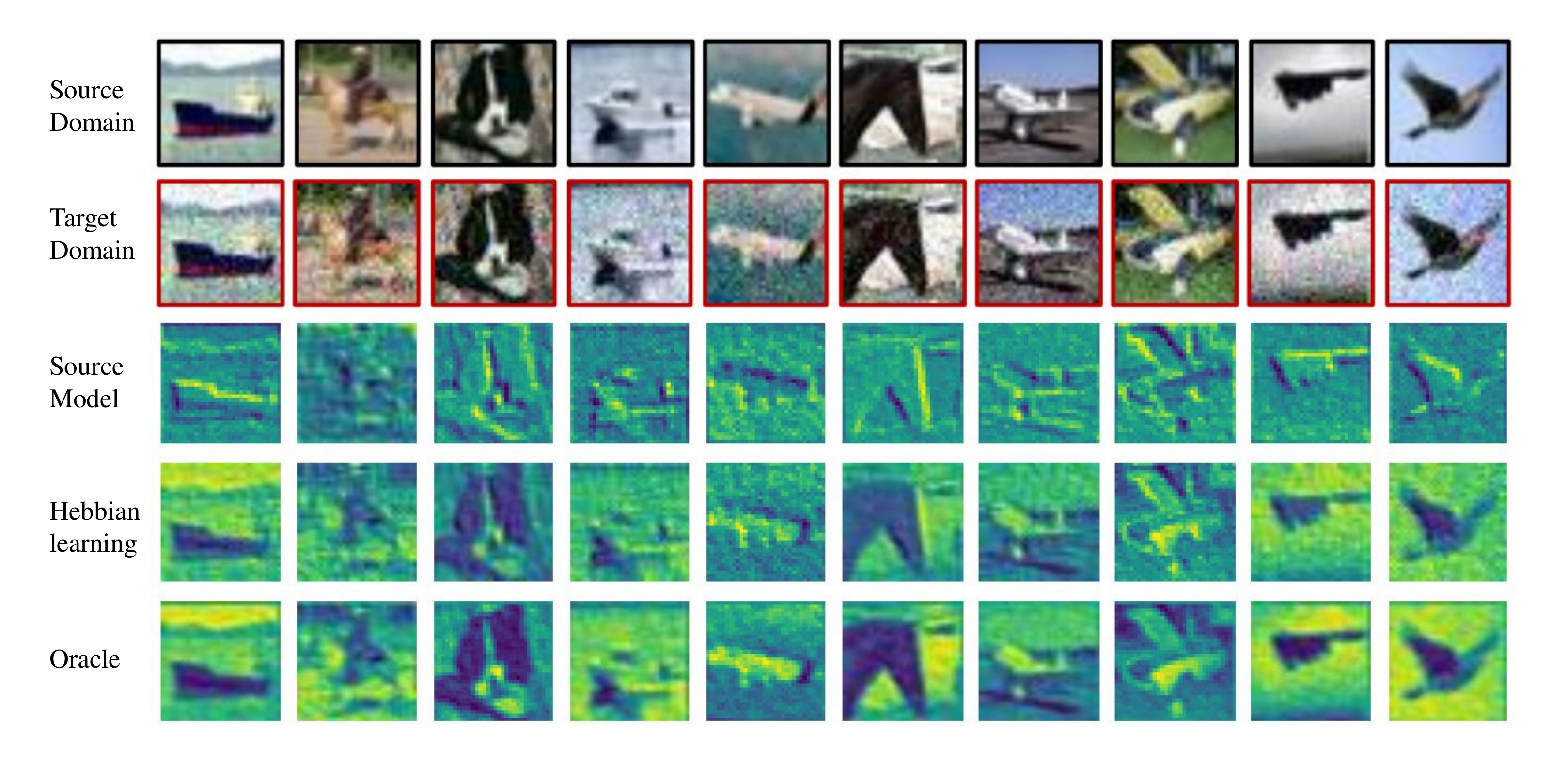}
\caption{The feature map visualization after the first convolution layer obtained by different learning method.}
\label{fig: motivation}
\end{figure}

\subsection{Hebbian Learning}
Hebbian learning aims to learn useful early layer representations without supervision based on local synaptic plasticity rules, which is able to generate early representations that are as good as those learned by end-to-end supervised training with back-propagation \cite{krotov2019unsupervised,pogodin2020kernelized}.
Drastically different from the current error back-propagation methods which require pseudo-labels or loss functions from the  top network layers, 
Hebbian learning is a pure feed-forward adaptation process and does not require feedback from the distant top network layers. 
The neuron responses are tuned based on a local synapse-change procedure 
and activated by competitive lateral inhibition rules \cite{krotov2019unsupervised}. 
Specifically, the local change of synapse strength during the learning process  is proportional to the activity of the pre-synaptic cell and to a function of the activity of the postsynaptic cell. 
It also introduces local lateral inhibition between neurons within a layer, where the synapses of hidden units with strong responses  are pushed toward the patterns that drive them, while those with weaker responses are pushed away from these patterns.

Existing literature has shown that early representations learned by Hebbian learning are as well as back-propagation and even more robust in testing~\cite{krotov2019unsupervised,pogodin2020kernelized,lagani2021hebbian}. Figure~\ref{fig: motivation} shows the feature maps
learned by different methods. The first row shows the original image. The second row shows the images in the target domain with significant image corruption. The third row shows the feature maps learned by the network model trained in the source domain
for these target-domain images. 
The fourth row shows the feature maps learned by our Hebbian learning method. The last row (``oracle") shows the feature maps learned with true labels. We can see that the unsupervised Hebbian learning is able to generate feature maps which are as good as those from supervised learning.

\subsection{Our Major Idea}
In this work, we observe that Hebbian learning, although provides a new and effective approach for unsupervised learning of early layer representation of the image, when directly applied to the network model, is not able to achieve satisfactory performance in fully test-time adaptation. First, the original hard decision for competitive learning is not suitable for fully test-time adaptation. Second, the Hebbian learning does not have an effective mechanism to consider external feedback, especially the feedback from the top network layers.
We observe that, biologically, the visual processing is realized through hierarchical models considering a bottom-up early representation learning for the sensory input, and a top-down feedback mechanism based on predictive coding \cite{rao1999predictive,friston2006free}. 

Motivated by this, in this work, we propose to develop a new approach, called \textit{neuro-modulated Hebbian learning (NHL)}, for fully  test-time adaptation. 
We first incorporate a soft decision rule into the feed-forward Hebbian learning to improve its competitive learning. 
Second, we learn a neuro-modulator to capture feedback from external responses, which controls which type of feature is consolidated and further processed to minimize the predictive error.
During inference, the source model is adapted by the proposed NHL rule for each mini-batch of testing samples during the inference process.
Experimental results on benchmark datasets demonstrate that our 
proposed method can significantly improve the 
adaptation performance of network models and outperforms existing state-of-the-art methods.

\subsection{Summary of Major Contributions}
To summarize, our major contributions include: (1) we identify that the major challenge in fully test-time adaptation lies in effective unsupervised learning of early layer representations, and explore neurobiology-inspired soft Hebbian learning for effective early layer representation learning and fully test-time adaptation.
(2) We develop a new neuro-modulated Hebbian learning method which combines unsupervised feed-forward Hebbian learning of early layer representation with a learned neuro-modulator to capture feedback from external responses. 
We analyze the optimal property of the proposed NHL algorithm based on free-energy principles~\cite{friston2006free,friston2009free}.
(3) We evaluate our proposed NHL method on  benchmark datasets for fully test-time adaptation, demonstrating its significant performance improvement over existing methods. 

\section{Related Work}
In this section, we review existing methods related to our work including test-time adaptation, source-free UDA, domain generalization, and unsupervised Hebbian learning.

\subsection{Test-time Adaptation}
Test-time adaptation aims to online adapt the trained model while testing the input samples in the target domain. Sun \etal \cite{DBLP:conf/icml/SunWLMEH20} proposed Test-time Training (TTT) by optimizing a self-supervised loss through a proxy task on the source before adapting to the target domain.
TTT++~\cite{liu2021ttt++} encouraged the test feature distribution to be close to the training one by matching the moments estimated online. It should be noted that this method requires specific training on the source data, which is not available in fully test-time adaptation scenarios \cite{wang2020tent}.
TENT~\cite{wang2020tent} fine-tuned the scale and bias parameters of batch normalization layers using an entropy minimization loss during the inference process.
DUA~\cite{mirza2022norm} adapted the statistics of batch normalization layer only on a tiny fraction of test data and augmented a small batch of target data to adapt the model.
Choi \etal \cite{choi2022improving} proposed a shift-agnostic weight regularization and an auxiliary task for the alignment between the source and target features.  Note that this method requires the source data for computing the source prototypes.
The continual test-time adaptation methods \cite{wang2022continual, marsden2022gradual} consider online TTA where target data is continually changing during inference. 
Instead of using parameters of the pre-trained model, Boudiaf \etal \cite{boudiaf2022parameter} only adapted the model's output by optimizing an objective function based on  Laplacian adjusted maximum-likelihood estimation. Besides image classification, test-time adaptation has been successfully applied in various machine learning tasks, such as scene deblurring \cite{chi2021test}, super-resolution \cite{shocher2018zero}, human pose estimation \cite{li2021test}, image segmentation~\cite{hu2021fully}, object detection \cite{mirza2022norm}, \etc. 

\subsection{Source-free Unsupervised Domain Adaptation}
Recently, source-free UDA has emerged as an important research topic. It aims to adapt the source models to unlabeled target domains without accessing the data from the source domain. Among them, SHOT~\cite{liang2020we} proposed to learn target-specific features based on an information maximization criteria and pseudo-label prediction. 3C-GAN~\cite{li2020model} generated target-style images by training a collaborative class conditional GAN module and using a clustering-based regularization scheme. G-SFDA~\cite{Yang_2021_ICCV} proposed a domain-specific attention to activate different feature channels for different domains. CPGA~\cite{qiu2021source} proposed to generate avatar prototypes instead of images via contrastive learning. HCL~\cite{huang2021model} designed historical contrastive instance discrimination and historical contrastive category discrimination to make up for the absence of source data. DIPE~\cite{Wang_2022_CVPR} explored the domain-invariant parameters of the model rather than attempting to learn domain-invariant representations.
SFDA-DE~\cite{ding2022source} aligned domains by estimating source class-conditioned feature distribution and minimizing a contrastive adaptation loss function.
Existing source-free methods are offline. They need to analyze the whole test dataset and update the model for a number of adaptation epochs. 

\subsection{Domain Generalization}
Domain generalization aims to train a model only on the source data that are generalizable to unseen target domains \cite{zhou2022domain}. Typical domain generalization methods attempt to align source domain distributions for domain-invariant representation learning \cite{muandet2013domain, motiian2017unified, zhao2020domain}. Methods based on data augmentation \cite{volpi2019addressing, qiao2020learning} for regularizing the training process have been studied. Ensemble learning methods \cite{ding2017deep, seo2020learning}  generate an ensemble of different model weights from different partitions  of the training data.
Recently, inference-time optimization without training the network model has been studied for domain generalization. Pandey \etal \cite{pandey2021generalization} used a generative model to project the target data onto the source-feature manifold with labels being preserved by solving an optimization problem during the inference stage.
Iwasawa \etal \cite{iwasawa2021test} proposed a back-propagation-free generalization method by computing distance to a pseudo-prototype representation. 

\subsection{Unsupervised Hebbian Learning}
In traditional end-to-end training, back-propagation is usually used to update the weights of deep neural networks. 
 It has been recognized that the learning rule of supervised end-to-end deep neural network training using back-propagation is much different from the learning rules of the early front-end neural processing in neurobiology.
In addition, supervised training of deep neural networks with back-propagation requires a large amount of labeled samples~\cite{lecun2015deep}.
In real biological neural networks, the neuron responses are tuned by a synapse-change procedure that is physically local.
Neurons activated by competition through lateral inhibition are a common connectivity pattern in the superficial cerebral cortex~\cite{douglas2004neuronal,binzegger2004quantitative}. 
Competition suppresses the activity of weakly activated neurons and emphasizes strong ones, which is known as competitive learning or “winner-takes-all (WTA)” learning~\cite{rumelhart1985feature,rutishauser2011collective}. Motivated by Hebb’s idea~\cite{hebb2005organization}, several biological plausibility learning rules have been proposed, where changes of the synapse strength depend only on the activities of the pre- and post-synaptic neurons. The Oja's rule~\cite{oja1982simplified} proposed a linear neuron model with constrained Hebbian synaptic modification and derived a new unconstrained learning method. 
Krotov \etal~\cite{krotov2019unsupervised} proposed a family of learning rules which conceptually have biological plausibility and allow learning early representations that are as good as those learned by the end-to-end supervised training with back-propagation. 
Pogodin \etal~\cite{pogodin2020kernelized} presents a family of learning rules which use pre- and post-synaptic firing rates and a global teaching signal. They perform almost the same as the back-propagation method.

\section{Method}
In this section, we present our method of neuro-modulated Hebbian learning for fully  test-time adaptation.

\subsection{Problem Formulation}
Suppose that a model ${m}(\theta)$ with parameters $\theta$ has been successfully trained on the source data $\{X_s\}$ with labels $\{Y_s\}$. It  consists of a feature extractor $\mathbf{F}$ and a classifier $\mathbf{C}$. During fully test-time adaptation, we are given the target data $\{X_t\}$ with unknown labels $\{Y_t\}$. Our goal is to adapt the trained model ${m}$ in an unsupervised manner during testing. Given a sequence of input sample batches $\{B_1, B_2, ..., B_n\}$, the $i$-th adaptation of the network model can only rely on the $i$-th batch of test samples ${B}_i$. 

We consider this transfer learning on a new target domain as an active inference process. The pre-trained model ${m}$ is considered as a prior. Thus, predicting the labels for target samples becomes a Bayesian posterior  $q_m(y| B_1)$ for the first batch, given the prior model $p_m(y)$. After the first adaptation batch, the prior model is updated to $m_1$. This Bayesian inference process is repeated for all subsequent batches and produces a final posterior estimation
\begin{equation}
    q_{m}(y|X_t)=q_{m_{n-1}}(y |B_n),
\end{equation}
where ${m}_{n-1}$ denotes the prior model after continual adaptation on the first $n-1$ batches. 

\subsection{Overview of Our NHL Method}
\begin{figure}[!t]
\setlength{\belowcaptionskip}{-0.45cm}
\centering
\includegraphics[width = 0.46\textwidth]{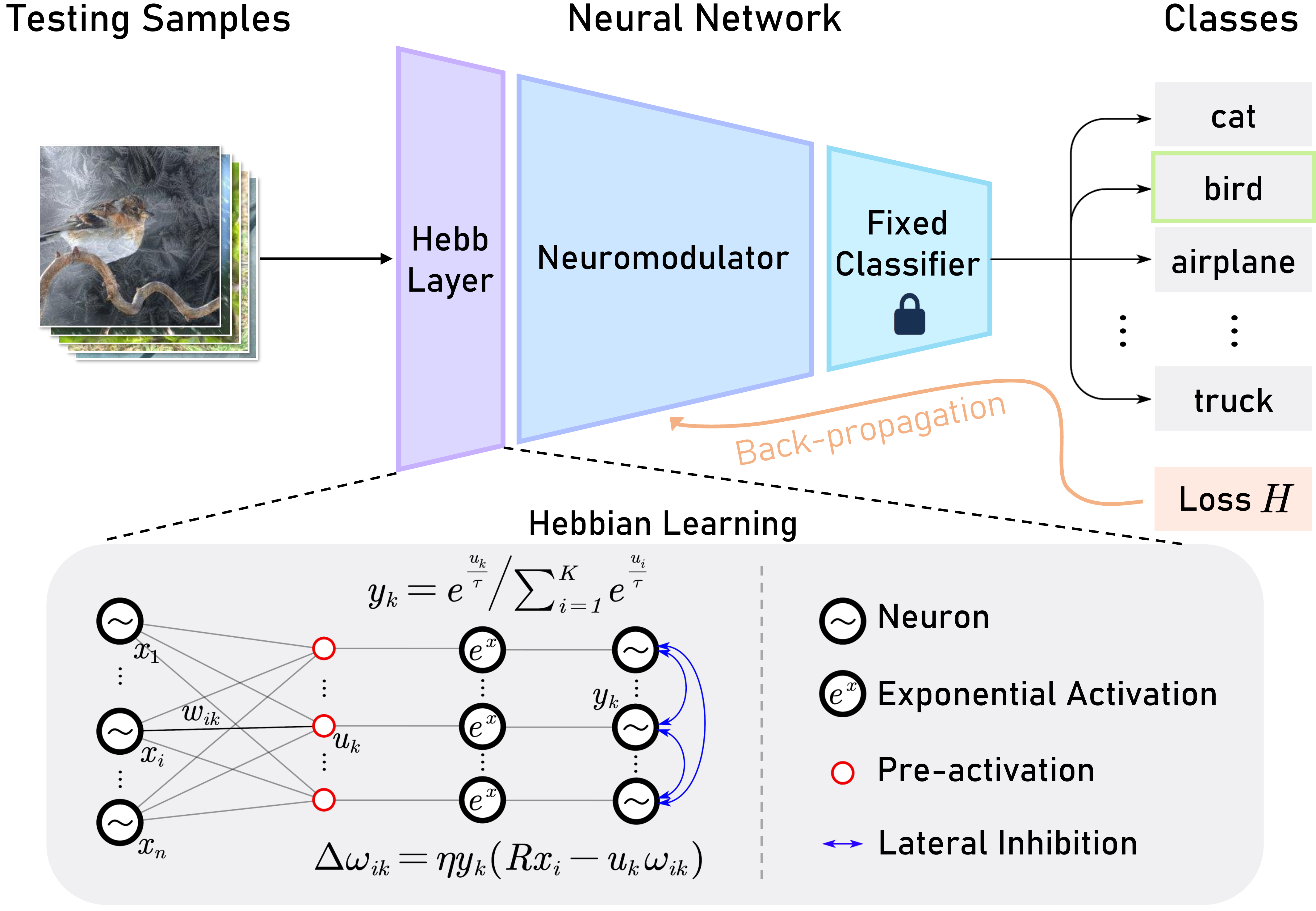}
\caption{An overview of the proposed NHL method. During inference, the first convolution layer of the source model is fine-tuned by Hebbian learning rule and the neuromodulator is further fine-tuned by the entropy loss before making a prediction given each mini-batch testing sample. The lock symbol means the classifier is fixed in the test-time adaptation process.}
\label{fig: overview}
\end{figure}

As illustrated in Figure~\ref{fig: overview}, 
our proposed neuro-modulated Hebbian learning consists of two major components: the feed-forward soft Hebbian learning layer and the neuro-modulator. 
The soft Hebbian learning layer aims to learn useful early layer representations without supervision based on local synaptic plasticity and soft competitive learning rules. 
It is able to generate early representations which are as good as those learnt by end-to-end supervised training with labeled samples and back-propagation. 
During our experiments, we find that this soft Hebbian learning layer can significantly improve the performance of the network model in the target domain. However, we recognize that the feed-forward Hebbian learning only is not able to achieve competitive  performance as the current state-of-the-art methods for fully test-time adaptation. We find that it lacks the capability to effectively respond to external stimulus, specifically the feedback from the network output layer. 

To address this issue, we propose to design a neuro-modulator layer, an intermediate layer or an interface between the soft Hebbian learning layer and the classifier module of the network. This neuro-modulator layer is updated using back-propagation with the entropy loss being evaluated at the network output. 
It serves as the bridge between two different learning approaches: the feed-forward Hebbian learning and the original error back-propagation. 
From the ablation studies summarized in Table \ref{table: ablation}, we can see that both algorithm components, the feed-forward soft Hebbian learning and the neuro-modulator are contributing significantly to the overall performance improvement.

The proposed NHL method can be also formulated  by the free-energy principle in cognitive science~\cite{friston2006free,friston2009free,rao1999predictive}. 
A hierarchical predictive coding model was proposed by Rao and Ballard~\cite{rao1999predictive} to learn a hierarchical internal model for human perception by maximizing the posterior probability of generating the observed data. This can be realized by a concurrent process of prediction through 
a bottom-up feed-forward generation process (such as our Hebbian learning layer) combined with a top-down feedback-based optimization process (such as our neuro-modulator). 
Specifically, given the sensory input $x_t$, assume $x_t$ is generated by environmental causes $\vartheta$, denoted as $p(x_t) = p(x_t, \vartheta)$. The free-energy principle states that the brain encodes the recognition density over sensory causes~\cite{friston2009free}. Mathematically, it optimizes a generative probabilistic mixture as $q_1(x_t):=q_1(x_t, \vartheta)$. It has been demonstrated that this process can be achieved by a Hebbian-like learning approach \cite{nessler2013bayesian,moraitis2022softhebb}. 

On the other hand, to approximate the true distribution $p(y |X_t )$ by a posterior approximation $q_2(y):=q_2(y | X_t)$, one can consider the similarity between these two distributions measured by the following Kullback-Leibler (KL) divergence
\begin{equation}\label{eq_KL}
    \mathrm{KL}[q_2(y) || p(y |X_t)] = \int q_2(y) \log \frac{q_2(y)}{p(y | X_t)}dy.
\end{equation}
According to the analysis in Bogacz \cite{bogacz2017tutorial},
this minimization of KL-divergence can be converted to maximization of the \textit{free-energy} $F$ defined as:
\begin{equation}\label{eq_free energy}
    F = \int q_2(y) \log \frac{p(y, X_t)}{q_2(y)}dy.
\end{equation}
The maximization of the free-energy $F$ can be solved by an active inference process if we consider the adaptation for each batch of test samples as a decision-making step \cite{friston2008hierarchical}. This leads  to a feedback-based optimization, just like the neuro-modulation as proposed in our NHL method. In the following sections, we explain the two major components of our NHL method in more detail.

\subsection{Feed-Forward Soft Hebbian Learning}
As discussed above, the early feed-forward layer aims to learn useful layer representations of the input without any supervision. An approximate solution to this problem can be reduced to finding the first principal component of the input data \cite{oja1982simplified}, known as Oja's rule and its variations. The Oja's rule is itself a variation of the Hebbian rule, plus a normalization condition. In Oja's rule, the plasticity defined for a synaptic weight $w_{ik}$ which connects a pre-synaptic neuron $i$ with input $x_i$ to a postsynaptic neuron $k$ with activation $y_k$ is: 
\begin{equation}\label{eq:oja}
    \Delta w_{ik} = \eta y_k (x_i - y_k w_{ik}).
\end{equation}
These weights are updated based on a plasticity rule \cite{krotov2019unsupervised}. We recognize that this process lacks the ability to detect different features of the data by different hidden units in the network. To address this problem, we propose to identify a good set of weights, by considering a generative model to optimize the distribution similarity to the input data. 

Specifically, we assume that the  target-domain input samples $\{X_t\}$ are generated by hidden causes $\vartheta$ with distribution $p_t(x)$. We define an approximation to $p_t(x)$ by $q(x | \boldsymbol{w})$ conditioned on the weights $\boldsymbol{w}$. Moreover, we use a mixture of exponential functions to define the probability:
\begin{equation}\label{eq:qdefine}
    q(x | \boldsymbol{w}_k) = \exp (<\boldsymbol{w}_k, x>)/N,
\end{equation}
where $\boldsymbol{w}_k$ is the weight vector corresponds to the post-synaptic neuron $k$, and $N$ is a normalization factor to ensure that $q$ be a probabilistic measure.
In the above generative model, the objective function is given by the Kullback-Leibler (KL) divergence $\mathrm{KL}[p_t(x)|q(x | \boldsymbol{w})]$. It can be shown that the optimal parameter vector that optimizes the KL-divergence is proportional to the mean of the input distribution~\cite{nessler2013bayesian}.

On the other hand, \eqref{eq:qdefine} corresponds to neural interpretation as the activation function with normalized weights $\boldsymbol{w_k}/R$, where $R$ is the norm of the weight vector $\boldsymbol{w_k}$. 
In our soft Hebbian learning, for the network layer with $K$ neurons, we define the output of the $k$-th neuron to be:
\begin{equation}
\begin{aligned}
    y_k = {e^{\frac{u_k}{\tau}}}/\left({\sum_{i=1}^{K} e^{\frac{u_i}{\tau}}}\right),
\end{aligned}
\end{equation}
where $u_k$ is the $k$-th neuron’s weighted input, \ie $u_k = w_{ik}\cdot x_i$. $\tau$ is a temperature-scaling hyper-parameter. This leads to a new soft Hebbian plasticity rule:
\begin{equation}\label{eq:hebbian rule}
\begin{aligned}
    \Delta w_{ik} = \eta y_k (Rx_i - u_k w_{ik}),
\end{aligned}
\end{equation}
where $\eta$ is the learning rate. It can be shown that, using the plasticity rule in \eqref{eq:hebbian rule} to update the weights, they can converge to the equilibrium which lies in the sphere of radius $R$. To see this, we rewrite \eqref{eq:hebbian rule} as a differential equation:
\begin{equation}
    \tau \Delta w_{ik} = \tau \frac{dw_{ik}}{dt} = \tau\eta y_k (Rx_i - u_k w_{ik}),
\end{equation}
where the constant $\tau$ defines the time scale of the learning dynamics. The derivative of the norm of the weight vector is:
\begin{equation}
\begin{aligned}
    &\frac{d\parallel\boldsymbol{w}_{k}\parallel}{dt} = \frac{2\boldsymbol{w}_{k}}{\tau} \eta y_k (Rx - u_k \boldsymbol{w}_{k})\\
    &= \frac{2\eta}{\tau} \boldsymbol{w}_{k} y_k (Rx - \boldsymbol{w}_{k} x \boldsymbol{w}_{k})
    = \frac{2\eta}{\tau} u_k y_k (R - \parallel\boldsymbol{w}_{k}\parallel). \nonumber
\end{aligned}
\end{equation}
With this, we can see that the norm of the weight vector convergence to a sphere of radius $R$. This is because the norm of the weights decreases if $\|\boldsymbol{w}_k\| > R$ and increases if $\|\boldsymbol{w}_k\| < R$.
More details are provided in the Supplemental Materials.

\subsection{The Neuro-Modulation Layer}
From the ablation studies in Section~\ref{sec:experiments}, we can see that the 
above soft Hebbian learning alone does not automatically lead to a perfect posterior estimation for $p(y | X_t)$. It cannot achieve the state-of-the-art performance for fully test-time adaptation.
This is because it does not have an effective mechanism to consider external feedback, especially the feedback from the top network layers.
Our proposed solution is to incorporate  one or more modulating layers to steer the updates of weights to the desired outcome \cite{lillicrap2020backpropagation, gerstner2018eligibility}. This so-called neuro-modulator has been explored in neuroscience research \cite{o2012norepinephrine,picciotto2012acetylcholine,speranza2021dopamine,bougrova2022comparison}. Recent research \cite{lehr2022neuromodulator} shows that the level of neuro-modulation may change the process of synaptic consolidation, thus ultimately controlling which type of information is consolidated in the upper neural network. Unlike most existing neuromodulator-based learning, where the modulator factor is embedded inside the Hebbian rule, we consider such a neuromodulation process in a disentangled way derived from the free-energy principle. It serves as an interface between the feed-forward Hebbian layer and the top decision network. 

As defined in (\ref{eq_free energy}), the problem of minimizing the KL-divergence for $q_2(y)$ and its true posterior $p(y|X_t)$ can be formulated based on the free-energy principle:
\begin{equation}\label{eq:KL-free-energy}
    \mathrm{KL}[q_2(y) || p(y|X_t)] = -F + \log P_m(X_t),
\end{equation}
where $P_m(X_t) := \int p(y)p_m(X_t|y)dy$ is the normalization term. Note that this term does not depend on $q_2(y)$. Therefore, minimizing the KL-divergence is reduced to maximizing $F$. As a result, this reduces to the minimization of the likelihood function $\int p_{m_{i-1}}(y, \boldsymbol{B}_i)dy $, given the labels $y$ of batch data $\boldsymbol{B}_i$. 
Note that, during the fully test-time adaptation process, sample labels are not available. Instead, we minimize the  entropy of $y$ for each batch $\boldsymbol{B}_i$, based on an approximate posterior $q_{m_{i-1}}(y | \boldsymbol{B}_i)$. Thus, the loss function to be optimized during the training of the neuro-modulation layer becomes
\begin{align}\label{eq:entropy loss}
    &\mathop{\arg \min}_{w_n}\  H_{m_{i-1}}(y | \boldsymbol{B}_i) = \notag \\
    &\mathop{\arg \max}_{w_n}\sum q_{m_{i-1}}(y | \boldsymbol{B}_i) \log q_{m_{i-1}}(y | \boldsymbol{B}_i), 
\end{align}
where $w_n$ is the weights in the layer implementing neuro-modulator. As in existing deep neural network training, we can use gradient descent to optimize these weights.

\begin{table*}[!ht]
\setlength{\abovecaptionskip}{0cm}
\setlength{\belowcaptionskip}{-0.2cm}
\begin{center}
\caption{Top-1 Classification Error (\%) for each corruption in \textbf{CIFAR-10C} at the highest severity (Level 5).  For TTT, TENT, and DUA, we use the \textbf{ResNet-26} (top), \textbf{WRN-28-10} (middle) and \textbf{WRN-40-2} (bottom) from their official implementation. Smallest error is shown in bold.}
\label{table: CIFAR-10C}
\resizebox{\linewidth}{!}{
\begin{tabular}{l|ccccccccccccccc|r}
\toprule
Methods & gaus & shot & impul & defcs & gls & mtn & zm & snw & frst & fg & brt & cnt & els & px & jpg & Avg.\\			
\midrule
Source & 67.7 & 63.1 & 69.9 & 55.3 & 56.6 & 42.2 & 50.1 & 31.6 & 46.3 & 39.1 & 17.1 & 74.6 & 34.2 & 57.9 & 31.7 & 49.2\\
TTT~\cite{DBLP:conf/icml/SunWLMEH20} & 45.6 & 41.8 & 50.0 & 21.8 & 46.1 & 23.0 & 23.9 & 29.9 & 30.0 & 25.1 & {\bf12.2} & 23.9 & {\bf22.6} & 47.2 & {\bf27.2} & 31.4\\
NORM~\cite{schneider2020improving} & 44.6 & 43.7 & 49.1 & 29.4 & 45.2 & 26.2 & 26.9 & 25.8 & 27.9 & 23.8 & 18.3 & 34.3 & 29.3 & 37.0 & 32.5 & 32.9\\
TENT~\cite{wang2020tent} & 39.4 & 38.8 & 47.9 & 19.9 & 45.0 & 23.2 & 20.6 & 28.1 & 32.1 & 24.5 & 16.1 & 26.7 & 32.4 & 30.6 & 35.5 & 30.7 \\
DUA~\cite{mirza2022norm} & 34.9 & 32.6 & 42.2 & 18.7 & {\bf40.2} & 24.0 & 18.4 & {\bf23.9} & {\bf24.0} & 20.9 & 12.3 & 27.1 & 27.2 & 26.2 & 28.7 & 26.8 \\ \rowcolor{gray!20}
\textbf{Ours} & {\bf33.2} & {\bf30.6} & {\bf38.2} & {\bf17.7} & 41.2 & {\bf20.8} & {\bf17.4} & 24.0 & 27.2 & {\bf20.4} & 13.5 & {\bf21.1} & 28.4 & {\bf23.7} & 28.9 & \textbf{25.8}\\
\noalign{\smallskip}
\hline
\noalign{\smallskip}
Source & 72.3& 65.7& 72.9& 46.9	& 54.3& 34.8& 42.0& 25.1& 41.3& 26.0& 9.3	& 46.7& 26.6& 58.5& 30.3& 43.5\\
NORM~\cite{schneider2020improving} & 28.1& 26.1& 36.3& 12.8& 35.3& 14.2& 12.1& 17.3& 17.4& 15.3& 8.4& 12.6& 23.8& 19.7& 27.3 & 20.4\\
TENT~\cite{wang2020tent} & 24.8	& 23.5& 33.0	& 12.0& 31.8& 13.7& {\bf10.8} & 15.9& 16.2& 13.7& 7.9& 12.1& 22.0& 17.3& 24.2& 18.6\\
DUA~\cite{mirza2022norm} & 27.4 & 24.6 & 35.3 & 13.1 & 34.9 & 14.6 & 11.6& 16.8 & 17.5 & 13.1 & {\bf7.6} & 14.1 & 22.7 & 19.3 & 26.2 & 19.9\\ \rowcolor{gray!20}
\textbf{Ours} & {\bf23.6} & {\bf21.4} & {\bf30.9} & {\bf11.0} & {\bf31.1} & {\bf13.0} & 10.9 & {\bf14.2} & {\bf15.5} & {\bf13.0} & 8.0 & {\bf10.3} & {\bf21.8} & {\bf16.7} & {\bf22.4} & \textbf{17.6}\\
\noalign{\smallskip}
\hline
\noalign{\smallskip}
Source & 28.8 & 22.9 & 26.2 & 9.5 & 20.6 & 10.6 & 9.3 & 14.2 & 15.3 & 17.5 & 7.6 & 20.9 & 14.7 & 41.3 & 14.7 & 18.3\\
NORM~\cite{schneider2020improving} & 18.7	& 16.4	& 22.3	& 9.1 & 22.1 & 10.5	& 9.7 & 13.0 & 13.2 & 15.4 & 7.8 & 12.0 & 16.4 & 15.1	& 17.6 & 14.6\\
TENT~\cite{wang2020tent} & 15.7 & 13.2 & 18.8 & 7.9 & 18.1 & 9.0 & 8.0 & 10.4 & 10.8 & 12.4 & 6.7 & 10.0 & 14.0 & 11.4 & 14.8 & 12.1\\
DUA~\cite{mirza2022norm} & 15.4 & 13.4 & 17.3 & 8.0 & 18.0 & 9.1 & {\bf7.7} & 10.8 & 10.8 & 12.1 & 6.6 & 10.9 & 13.6 & 13.0 & 14.3 & 12.1\\ \rowcolor{gray!20} 
\textbf{Ours} & {\bf13.4} & {\bf12.3} & {\bf15.0} & {\bf7.5} & {\bf16.0} & {\bf8.7} & {\bf7.7} & {\bf9.1} & {\bf9.6} & {\bf10.1} & {\bf6.4} & {\bf8.2} & {\bf13.3} & {\bf9.3} & {\bf13.3} & {\bf10.7} \\ 
\bottomrule
\end{tabular}
}
\end{center}
\end{table*}

\begin{table*}[htbp]
\setlength{\abovecaptionskip}{0cm}
\setlength{\belowcaptionskip}{-0.2cm}
\begin{center}
\caption{Top-1 Classification Error (\%) for each corruption in \textbf{CIFAR-100C} at the highest severity (Level 5).}
\label{table:CIFAR-100C}
\resizebox{\linewidth}{!}{
\begin{tabular}{l|ccccccccccccccc|r}
\toprule
Methods & gaus & shot & impul & defcs & gls & mtn & zm & snw & frst & fg & brt & cnt & els & px & jpg & Avg.\\			
\midrule
Source & 65.7 & 60.1 & 59.1 & 32.0 & 51.0 & 33.6 & 32.4 & 41.4 & 45.2 & 51.4 & 31.6 & 55.5 & 40.3 & 59.7 & 42.4 & 46.7\\
NORM~\cite{schneider2020improving} & 44.7 & 44.2 & 47.4 & 32.4 & 46.4 & 32.9 & 33.0 & 39.0 & 38.4 & 45.3 & 30.2 & 36.6 & 40.6 & 37.2 & 44.2 & 39.5\\
TENT~\cite{wang2020tent} & 40.3 & 39.9 & 41.8 & 29.8 & 42.3 & 31.0 & 30.0 & 34.5 & 35.2 & 39.5 & 28.0 & 33.9 & 38.4 & 33.4 & 41.4 & 36.0\\
DUA~\cite{mirza2022norm} & 42.2 & 40.9 & 41.0 & 30.5 & 44.8 & 32.2 & 29.9 & 38.9 & 37.2 & 43.6 & 29.5 & 39.2 & 39.0 & 35.3 & 41.2 & 37.6 \\ \rowcolor{gray!20}
\textbf{Ours} & \textbf{38.4} & \textbf{37.1} & \textbf{36.2} & \textbf{28.4} & \textbf{41.0} & \textbf{29.3} & \textbf{29.7} &\textbf{32.2} & \textbf{33.1} & \textbf{36.1} & \textbf{26.4} & \textbf{30.9} & \textbf{36.2} & \textbf{30.8} & \textbf{38.3} & \textbf{33.6} \\ 
\bottomrule
\end{tabular}
}
\end{center}
\end{table*}

\begin{table*}[!htbp]
\setlength{\abovecaptionskip}{0cm}
\setlength{\belowcaptionskip}{-0.45cm}
\begin{center}
\caption{Top-1 Classification Error (\%) for each corruption in \textbf{ImageNet-C} at the highest severity (Level 5).}
\label{table:ImageNet-C}
\resizebox{\linewidth}{!}{
\begin{tabular}{l|ccccccccccccccc|r}
\toprule
Methods & gaus & shot & impul & defcs & gls & mtn & zm & snw & frst & fg & brt & cnt & els & px & jpg & Avg.\\			
\midrule
Source & 98.9 & 97.6 & 99.2 & 93.3 & 89.0 & 90.2 & 82.3 & 87.9 & 92.0 & 99.5 & 75.9 & 99.5 & 65.3 & 60.3 & 54.0 & 85.7\\
TTT~\cite{DBLP:conf/icml/SunWLMEH20} & 75.5 & 76.8 & 81.9 & 89.6 & 82.8 & 79.1 & 71.3 & 83.6 & 81.0 & 98.3 & 59.0 & 99.0 & 54.7 & 53.2 & {\bf49.6} & 75.7\\
NORM~\cite{schneider2020improving} & 60.2 & 60.7 & 59.8 & 76.6 & 68.7 & 67.4 & 64.2 & 64.6 & 66.2 & 74.7 & 57.0 & 88.8 & 55.8 & 53.0 & 52.3 & 64.7\\
TENT~\cite{wang2020tent} & 59.4 & 59.6 & 58.7 & {\bf72.5} & 66.1 & 64.9 & 62.1 & 62.2 & 64.9 & 68.6 & 55.2 & 97.9 & 54.5 & 52.1 & 51.7 & 62.7 \\
DUA~\cite{mirza2022norm} & 71.9 & 72.6 & 72.4 & 90.2 & 80.8 & 83.1 & 74.7 & 76.4 & 77.9 & 87.3 & 62.6 & 99.3 & 60.8 & 58.4 & 52.6 & 74.7\\ \rowcolor{gray!20}
\textbf{Ours} & {\bf56.7} & {\bf56.7} & {\bf56.6} & 73.3 & {\bf65.7} & {\bf61.0} & {\bf62.0} & {\bf58.6} & {\bf63.3} & {\bf63.9} & {\bf53.1} & {\bf77.5} & {\bf54.0} & {\bf52.0} & 51.5 	& {\bf60.4} \\ 
\bottomrule
\end{tabular}
}
\vspace{-10pt}
\end{center}
\end{table*}

\section{Experiments}
\label{sec:experiments}
In this section, we conduct experiments on multiple test-time adaptation benchmark datasets to evaluate the performance of our proposed NHL method.

\subsection{Benchmark Datasets}
We evaluate our method among the following popular benchmark datasets for Test-time adaptation. (1) \textbf{CIFAR10/100-C.} We choose CIFAR10/100~\cite{krizhevsky2009learning} with 10/100 classes, a source training set of $50,000$, and a target testing set CIFAR10/100~\cite{hendrycks2018benchmarking} of $10,000$ for small-size image experiments at an accessible scale.
(2) \textbf{ImageNet-C.} We choose the ImageNet~\cite{russakovsky2015imagenet} with $1,000$ classes, a source training set of $1.2$ million, and a validation set of $50,000$ for large-size image experiments.
It should be noted that we use a fixed target testing subset of the validation set with 5000 images on which all models are evaluated following the RobustBench protocol \cite{croce2021robustbench}.
(3) \textbf{SVHN$\rightarrow$MNIST/MNIST-M/USPS.} Following TENT for domain adaptation, we choose a SVHN~\cite{netzer2011reading} source model and transfer it to MNIST~\cite{lecun1998gradient} ($10,000$ test samples), MNIST-M~\cite{ganin2015unsupervised}  ($10,000$ test samples) and USPS~\cite{hull1994database}  ($2,007$ test samples), respectively.

\subsection{Comparison Methods}
We compare our method against the following fully test-time adaptation methods: (1) \textbf{Source:} the baseline model is trained only on the source data without any fine-tuning during the test process.
(2) \textbf{TTT}~\cite{DBLP:conf/icml/SunWLMEH20}: it adapts the feature extractor by optimizing a self-supervised loss through a proxy task. However, it requires training the same proxy task on the source domain.
(3) \textbf{NORM}~\cite{schneider2020improving}: this test-time normalization method updates the batch normalization~\cite{ioffe2015batch} statistics using the mini-batch samples during the test process.
(4) \textbf{TENT}~\cite{wang2020tent}: it fine-tunes scale and bias parameters of batch normalization layers using an entropy minimization loss during inference. 
(5) \textbf{DUA}~\cite{mirza2022norm}: it adapts the statistics of the batch normalization layer only on a tiny fraction of test data and augmented a small batch of target data to adapt the model.

\subsection{Implementation Details}
Following the official implementations of 
TTT\footnote{TTT: \href{https://github.com/yueatsprograms/ttt_cifar_release}{https://github.com/yueatsprograms/ttt\_cifar\_release}}, 
TENT\footnote{TENT: \href{https://github.com/DequanWang/tent}{https://github.com/DequanWang/tent}} and 
DUA\footnote{DUA: \href{https://github.com/jmiemirza/DUA}{https://github.com/jmiemirza/DUA}}, 
we use the ResNet-26~\cite{he2016deep}, Wide-ResNet-28-10~\cite{zagoruyko2016wide}, and Wide-ResNet-40-2~\cite{zagoruyko2016wide} as the backbone networks for the CIFAR10-C dataset. We use the Wide-ResNet-40-2 network for the CIFAR100-C dataset. For the ImageNet-C dataset, we use the ResNet-18~\cite{he2016deep} backbone network. We use the pre-trained model weights from the official implementations of TENT or DUA for all backbones based on the RobustBench protocol \cite{croce2021robustbench}.
For the digit recognition transfer tasks, we use the pre-trained model weights of SVHN from the \textit{pytorch-playground}~\cite{playground}.
For fair performance comparisons, all methods in each experimental condition share the same architecture and the same pre-trained model parameters. The batch size is set to 128.
More implementation details are provided in the Supplemental Materials.

\subsection{Performance Results}
\begin{figure}[!ht]
\setlength{\belowcaptionskip}{-0.45cm}
\centering
\includegraphics[width = 0.38\textwidth]{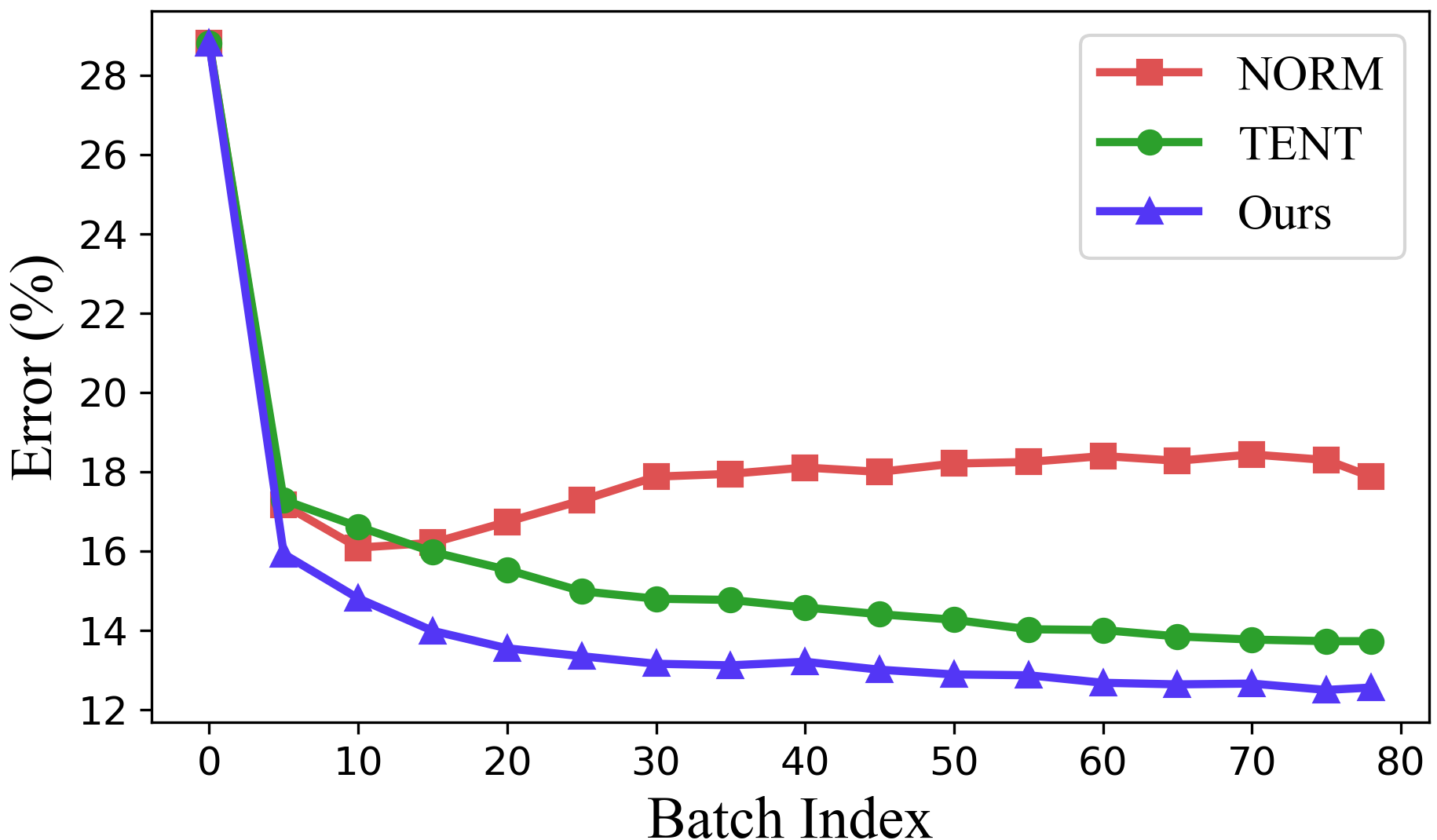}
\vspace{-5pt}
\caption{The mean adaptation error of the full test dataset on Gaussian noise of CIFAR-10C in the different adaptation stages.}
\label{fig: batch_error}
\end{figure}

The classification errors in the highest severity level corruption test datasets for test-time adaptation are reported in Tables~\ref{table: CIFAR-10C}, \ref{table:CIFAR-100C} and \ref{table:ImageNet-C}, with results of comparison methods directly cited from their original papers.
Table~\ref{table: CIFAR-10C} compares the classification error of our proposed method against recent test-time adaptation methods on the CIFAR-10C dataset.
Our method performs better than other baselines with the three backbones including ResNet-26, WRN-28-10, and WRN-40-2, indicating the effectiveness of the proposed test-time adaptation method.
Table~\ref{table:CIFAR-100C} shows the performance comparison results on the CIFAR-100C dataset. 
Very encouraging results are also obtained on the large-size complicated ImageNet-C dataset, as shown in Table~\ref{table:ImageNet-C}. 
The mean adaptation error of the full test dataset on Gaussian noise of CIFAR-10C in the different adaptation stages is shown in Figure~\ref{fig: batch_error}. We can see that our method outperforms the NORM and TENT method after the 5-th batch test-time adaptation. This implies our method can reduce error faster given few testing samples.
Results for lower severity levels of corruption are provided in Supplemental Materials.

Results for digit recognition from the SVHN to MNIST, MNIST-M, and USPS datasets are also reported in Table~\ref{table: SVHN}. All experiments use the same open-source source model. Note that 
the result for the TENT method is reproduced here since the pre-trained model for the TENT method was not provided and explained in its original paper. 
We can see that our method achieves the lowest average error when compared to other test-time adaptation methods.
The performance improvement is quite impressive.

\begin{table}[htbp]
\setlength{\abovecaptionskip}{0cm}
\setlength{\belowcaptionskip}{-0.15cm}
\begin{center}
\caption{Top-1 Classification Error (\%) for test-time adaptation on digit recognition. $^*$ means the implementation with a \textit{pytorch-playground}~\cite{playground} pre-trained source model by us.}
\label{table: SVHN}
\resizebox{0.45\textwidth}{!}{
\begin{tabular}{l|ccc|r}
\toprule
Methods & MNIST & MNIST-M & USPS & Avg.\\			
\midrule
NORM$^*$~\cite{schneider2020improving} & 39.6 &52.1 &41.4 &44.4\\
TENT$^*$~\cite{wang2020tent} & 45.8 &56.2 &48.3 &50.1\\ \rowcolor{gray!20}
Ours & \textbf{31.2} &\textbf{47.9} &\textbf{32.6} & \textbf{37.2}\\
\bottomrule
\end{tabular}
}
\end{center}
\vspace{-20pt}
\end{table}

\subsection{Further Performance Analysis}
\subsubsection{Ablation Study}
We conduct the ablation study with test-time adaptation tasks on the ImageNet-C dataset to investigate the contribution of our method. 
Hebbian learning alone does not automatically lead to a perfect posterior estimation for $p(y | X_t)$ due to the lack of global information communication.
From Table~\ref{table: ablation}, we can see that feed-forward soft Hebbian Learning plays a significant effect and fine-tuning the feedback neuro-modulator of Block 1/2 also improves the performance. The average error is increased when expanding the neuro-modulator to more blocks. It is because optimizing more parameters with a mini-batch of testing samples is becoming more challenging.

\begin{table}[htbp]
\setlength{\abovecaptionskip}{0.1cm}
\setlength{\belowcaptionskip}{-0.3cm}
\begin{center}
\caption{Ablation study on \textbf{ImageNet-C} based on ResNet-18 including 4 Blocks at the highest severity (Level 5). }
\label{table: ablation}
\resizebox{0.475\textwidth}{!}{
\begin{tabular}{l|c}
\toprule
Methods  & Avg. Error\\			
\midrule
Source (without adaptation) & 85.7\\
BP Conv1  & 85.0 \\
Hebbian Conv1  & 67.2 \\
Hebbian Conv1 + Neuromodulator (Block 1)	& 60.5 \\
Hebbian Conv1 + Neuromodulator (Block 1/2)	& {\bf60.4} \\ 
Hebbian Conv1 + Neuromodulator (Block 1/2/3)	& 61.2 \\
Hebbian Conv1 + Neuromodulator (Block 1/2/3/4)	& 64.8 \\
\bottomrule
\end{tabular}
}
\end{center}
\vspace{-20pt}
\end{table}

\subsubsection{Feature Visualization}
\begin{figure}[htbp]
\setlength{\belowcaptionskip}{-0.2cm}
\centering
\begin{subfigure}{0.23\textwidth}
  \centering
  \includegraphics[width=0.8\textwidth]{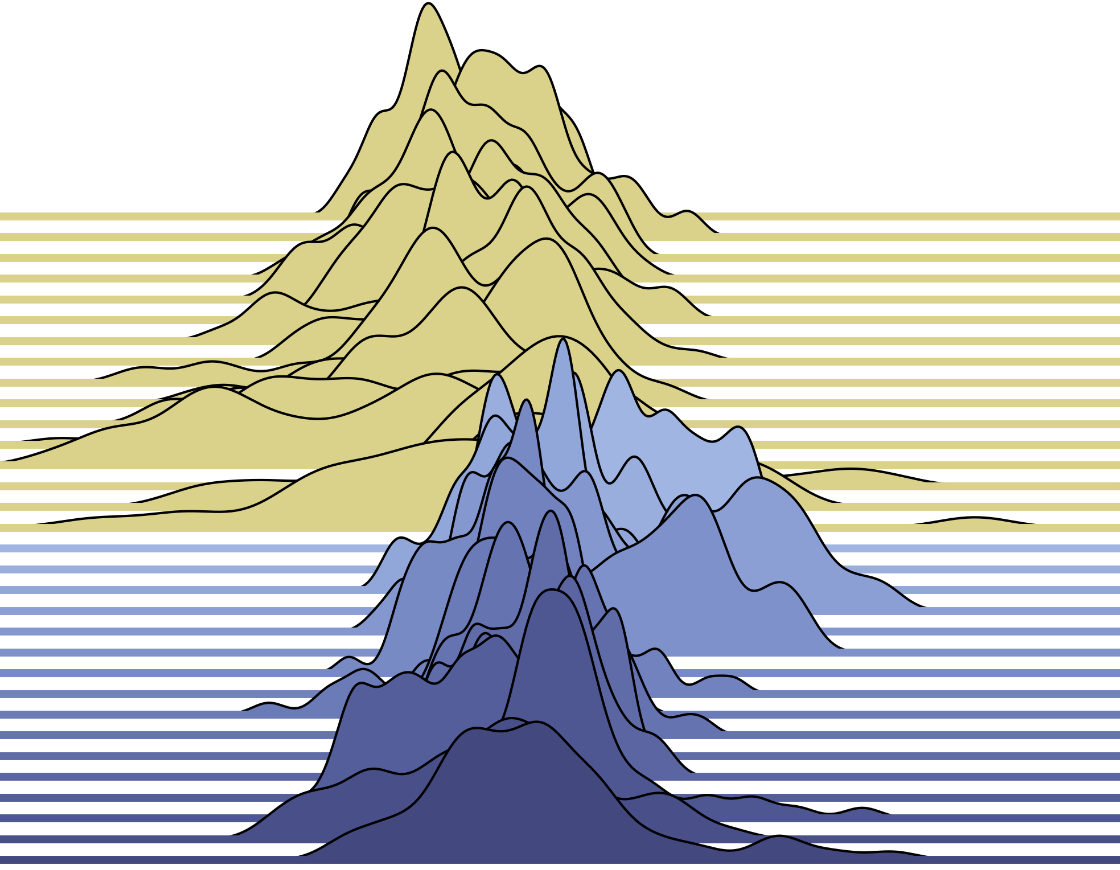}
  \caption{Source}
  \label{fig:source}
\end{subfigure}
\hskip -5pt
\begin{subfigure}{0.23\textwidth}
  \centering
  \includegraphics[width=0.8\textwidth]{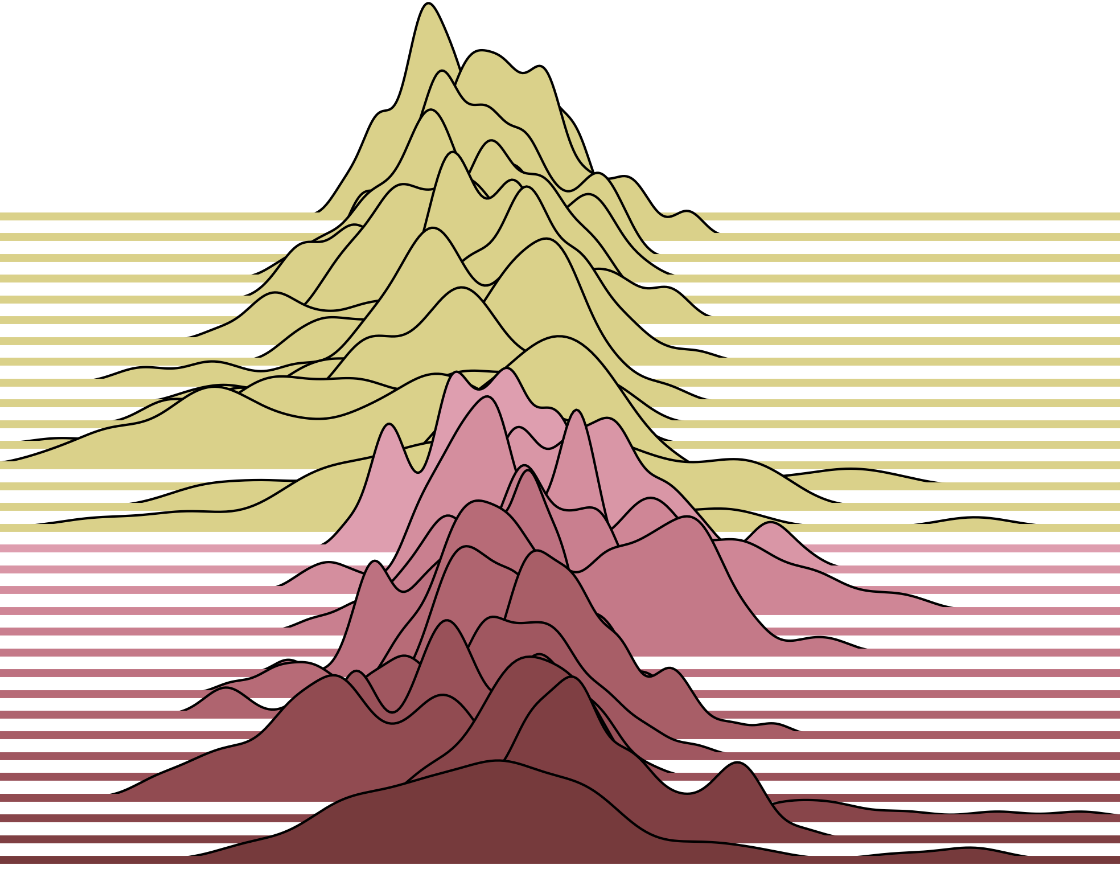}
  \caption{TENT}
  \label{fig:tent}
\end{subfigure}
\hskip -5pt
\begin{subfigure}{0.23\textwidth}
  \centering
  \includegraphics[width=0.8\textwidth]{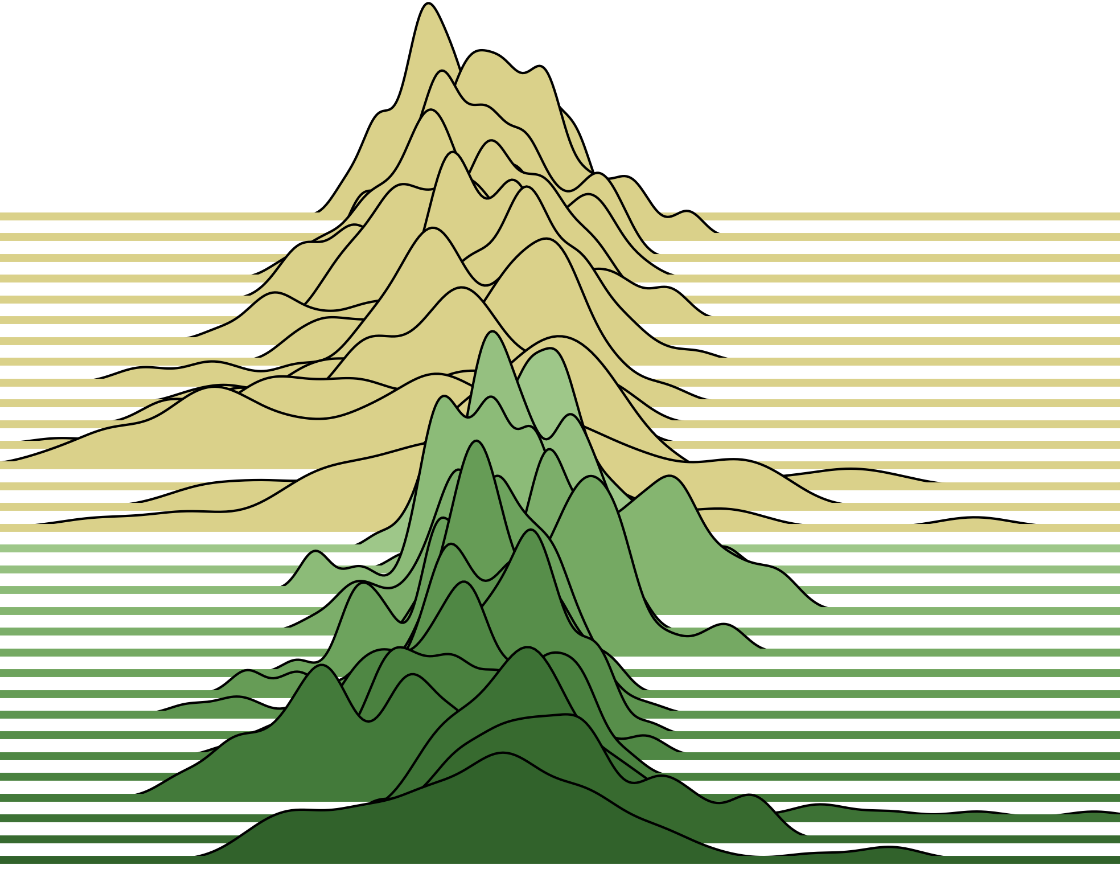}
  \caption{Ours}
  \label{fig:hebbian}
\end{subfigure}
\hskip -5pt
\begin{subfigure}{0.23\textwidth}
  \centering
  \includegraphics[width=0.8\textwidth]{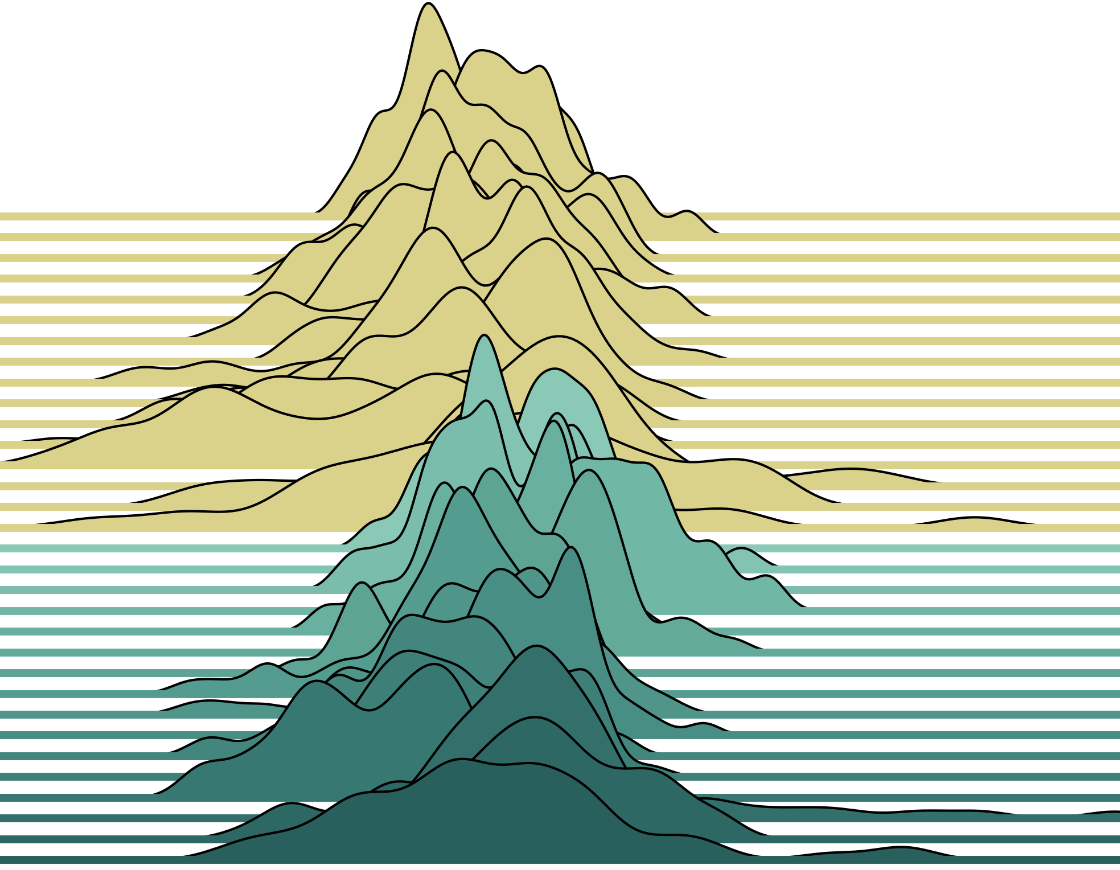}
  \caption{Oracle}
  \label{fig:oracle}
\end{subfigure}
\caption{Density plots of test-time adapted features distribution on CIFAR-10-C with Gaussian noise (front) and reference features without corruption (back with yellow color). Each horizontal axis corresponds to one channel. The height of each ridge indicates the number of features that take the same value.}
\label{fig:feature_visual}
\end{figure}

Figure~\ref{fig:feature_visual} compares the feature distributions on corrupted data obtained by different adaptation methods, including the source model with no adaptation, the TENT method, and our method. We also include the feature distribution for the supervised learning with labeled sample, denoted by ``Oracle". We can see that our method is able to learn features that are close to those obtained by the supervised learning.

\vspace{-15pt}

\section{Further Discussions}
Hebbian Learning by competition through lateral inhibition is a feed-forward process that has no gradient. If combining Hebbian Learning with back-propagation, it is limited to propagating the gradient through Hebbian layers to earlier gradient-based layers during training. Therefore, the Hebbian layers can only be designed in the early layers of models without back-propagation and gradient. In addition, although the Hebbian learning rule is commonly used for long-term reinforcement, the Hebb principle does not cover all forms of long-term synaptic plasticity.

\section{Conclusion}
In this work, we take inspiration from the biological neuromodulation process to construct a novel neuro-modulated Hebbian Learning (NHL) framework.
With the unsupervised feed-forward soft Hebbian learning
being combined with a learned neuro-modulator to capture
feedback from external responses, the source model can be
effectively adapted during the testing process. 
Experimental results on benchmark datasets demonstrate that our proposed method can significantly reduce the testing error for image classification with corruption, and reaches new state-of-the-art performance.

{\small
\bibliographystyle{ieee_fullname}
\bibliography{egbib}
}

\end{document}